\def\year{2020}\relax
\documentclass[letterpaper]{article} 
\usepackage{aaai20}  
\usepackage{times}  
\usepackage{helvet} 
\usepackage{courier}  
\usepackage[hyphens]{url}  
\usepackage{graphicx} 
\usepackage{graphicx}  
\usepackage{booktabs}
\usepackage[export]{adjustbox}
\usepackage{makecell,subcaption, caption}

\title{Learning to Emphasize:\\Dataset and Shared Task Models for Selecting Emphasis in Presentation Slides}

\author{Amirreza Shirani$^{\dagger}$, 
Giai Tran$^{\dagger}$,
Hieu Trinh$^{\dagger}$,
Franck Dernoncourt$^{\ddagger}$,\\
\Large\textbf{Nedim Lipka$^{\ddagger}$,
Paul Asente$^{\ddagger}$,
Jose Echevarria$^{\ddagger}$,
and Thamar Solorio$^{\dagger}$} \\
 	$^{\dagger}$University of Houston \qquad
 	$^{\ddagger}$Adobe Research \qquad \\
 	$^{\dagger}$\small{\texttt{\{ashirani,gltran,httrinh,tsolorio\}@uh.edu}} \\
 	$^{\ddagger}$\small{\texttt{\{franck.dernoncourt,lipka,asente,echevarr\}@adobe.com}} \\
 	}
 
\begin{document}

\maketitle

\begin{abstract}
Presentation slides have become a common addition to the teaching material. Emphasizing strong leading words in presentation slides can allow the audience to direct the eye to certain focal points instead of reading the entire slide, retaining the attention to the speaker during the presentation. Despite a large volume of studies on automatic slide generation, few studies have addressed the automation of design assistance during the creation process.
Motivated by this demand, we study the problem of Emphasis Selection (ES) in presentation slides, i.e., choosing candidates for emphasis, by introducing a new dataset containing presentation slides with a wide variety of topics, each is annotated with emphasis words in a crowdsourced setting. 
We evaluate a range of state-of-the-art models on this novel dataset by organizing a shared task and inviting multiple researchers to model emphasis in this new domain. We present the main findings and compare the results of these models, and by examining the challenges of the dataset, we provide different analysis components.
\end{abstract}

\section{Introduction}
\label{sec:intor}
The use of presentation slides has become so commonplace that researchers have developed resources meant to guide presenters in the design of effective slides \cite{alley2004rethinking,AlleyAndNeele:05,Jennings:09}. However, these guidelines cover only advice with respect to the overall style, such as colors and font size to ensure text is readable from a distance, as well as considerations with respect to graphical representations of content. 
However, users can benefit from complementary design recommendations during slide creation.
The result not only can be aesthetically appealing but can enhance the slides' content communication.
Moreover, related recommendations can potentially reduce the amount of time users spend in authoring and design. 

In this study, the main focus is on predicting emphasis words in presentation slides. Emphasis is the use of special formatting (e.g., \textbf{boldface} or \textit{italics}) to make a word or set of words stand out from the rest. Generally, word emphasis may use to express emotions, show contrast, capture a reader's interest, or clarify a message with a variety of applications on different platforms.
Specifically, in presentation slides, well-designed slides annotated with emphasis can significantly increase the audience's retention by guiding the audience into focusing on a few words \cite{alley2004rethinking}. Instead of reading the entire slide, the audience can read only the emphasized parts and retain their attention in the speaker. As an example of expected results, consider the slides shown in Figure \ref{fig:emph_ex} below. The slide on the top (a) is plain, and while the text is readable, the slide on the bottom (b) is easier for the audience to process.  

\begin{figure}[ht]
\centering
\begin{subfigure}{0.44\textwidth}
\includegraphics[width=1\linewidth, left]{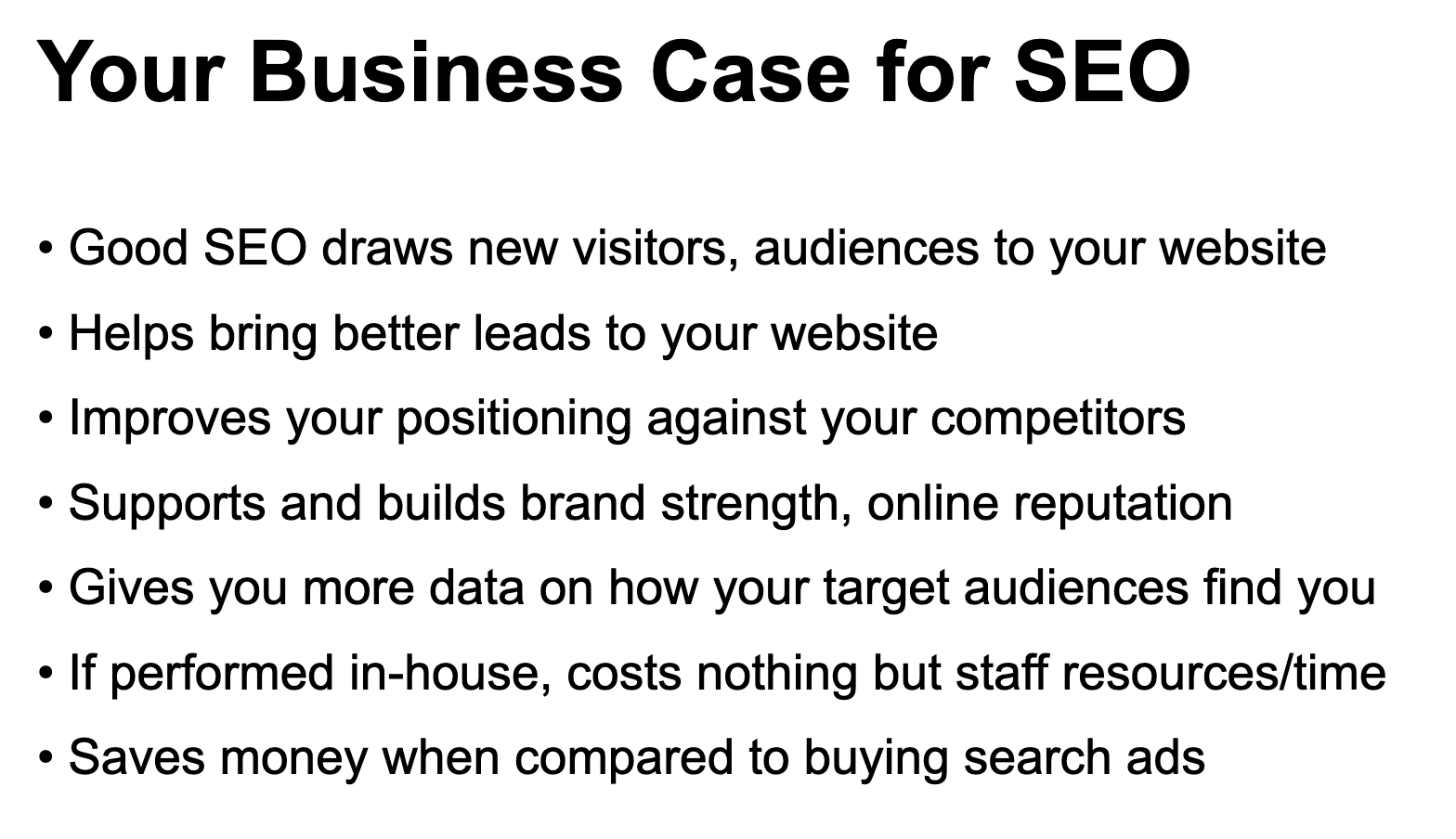} 
\caption{}
\label{fig:expa}
\end{subfigure}
\begin{subfigure}{0.44\textwidth}
\includegraphics[width=1\linewidth, right]{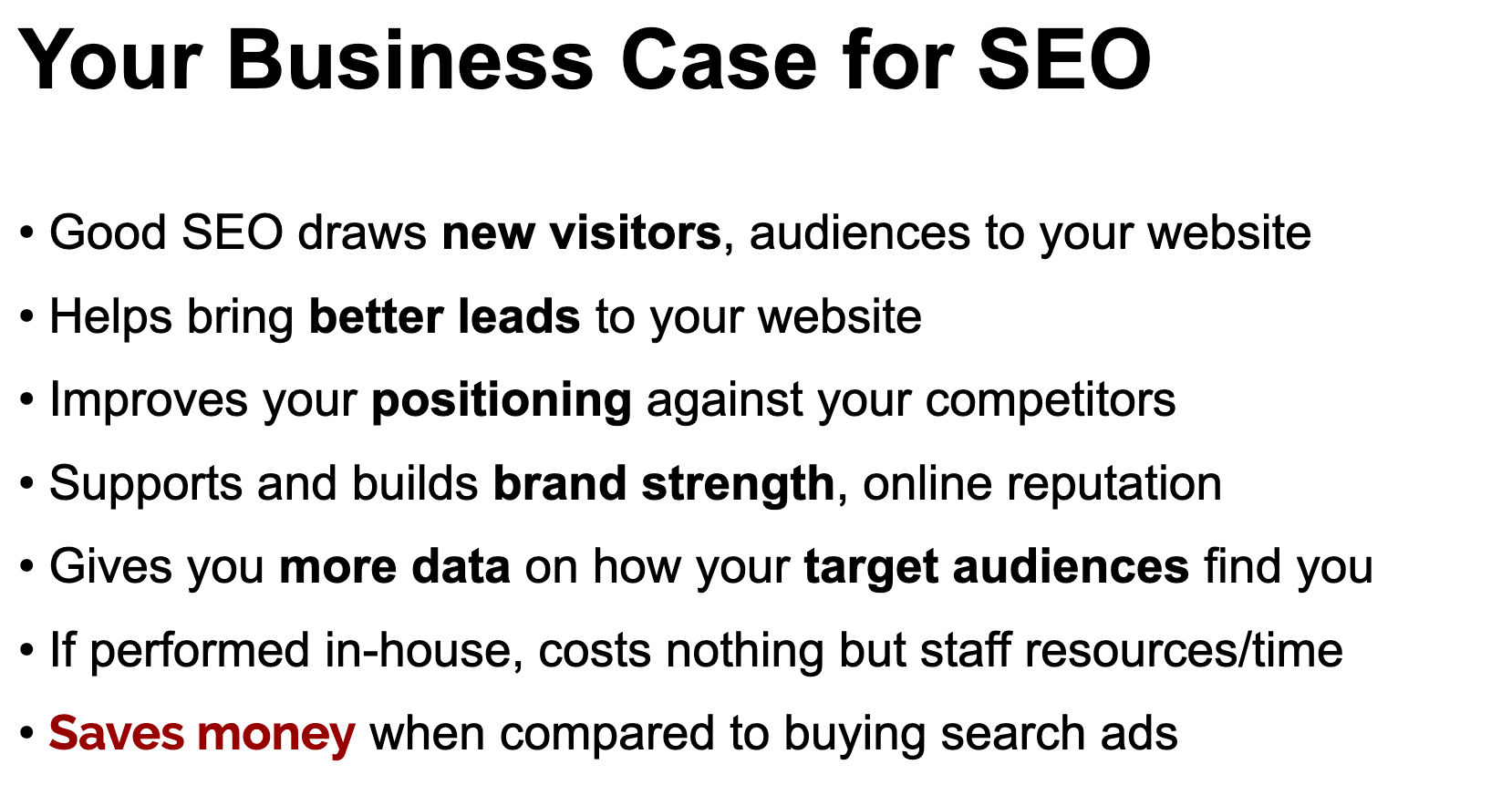}
\caption{}
\label{fig:expb}
\end{subfigure}
\caption{The figure shows side by side comparison of slides w/o emphasis. The slide on the top does not make use of any special formatting to highlight salient content while the one on the bottom does highlight key points.}
\label{fig:emph_ex}
\end{figure}
Emphasis selection (ES) task is initially introduced in \cite{shirani-etal-2019-learning} with the focus on social media short written text and has been featured and attracted attention in SemEval’s 2020, Emphasis Selection for Written Text in Visual Media Task \cite{shirani-etal-2020-semeval}.
In contrast to the previous works on ES, we introduce a new corpus focusing on presentation slides as well as automated approaches to predict emphasis on presentation slides with the goal of facilitating the understanding of the message and enhancing the visual appeal of the slides. This task is among the first to provide automated design assistance for presentation slides by relying on the content of the slides.

\paragraph{Task Characteristics} The emphasis selection task on presentation slides poses new challenges associated with the nature of the task: (1) Presentation slides can be found in different shapes and structures. Users depending on the usage, prefer to follow either traditional corporate presentation styles or modern styles with more visuals and less textual content. 
Due to their diverse usage, slides cover a wide range of topics from technical or legal presentations to non-technical ones such as children's illustrations. The requirement to generalize to different domains and cover a variety of topics poses new challenges and encourages the development of 
robust language understanding models. 
(2) To select emphasis words, we only rely on input text rather than other additional context from the user or the rest of the design. Therefore, to tackle the subjective nature of this task, we investigate models which by utilizing natural language understanding techniques aim to understand the most common interpretation of a slide page, so the right emphasis can be obtained automatically or interactively.


\paragraph{CAD21 Shared task} As part of this study, we organized a shared task\footnote{\url{https://competitions.codalab.org/competitions/27419}} and invited researchers to work on the new corpora. The goal of this shared task is to examine and introduce the state-of-the-art approaches to model emphasis in presentation slides. This shared task is part of the workshop on Content Authoring and Design (CAD21) at AAAI 2021.
In the following sections, we describe the top-performing systems, pointing out the insights gained from the task, the strength and limitations of these models, as well as future directions in this area.


\section{Related Work}
\label{sec:related}
In visual communication, a wide range of design components are typically used to increase the comprehension of content and to convey the author’s intent. 
Different authoring and graphic design applications perform automatic design assistance that include images and text in different forms and shapes. However, a majority of publicly available tools are mainly driven by some basic heuristics in assisting users during authoring. 
Recent works in the area started to employ AI-based models to assist users during authoring by recommending appropriate design components based on the content such as \cite{shirani-etal-2020-choose}.
Considering a wide range of applications and its unique challenges, this interdisciplinary area hasn't been fully studied and has little cross-disciplinary collaboration.

Many studies on prior works explored the automatic generation of presentation slides from documents such as scientific articles \cite{BeamerInvestigating,SidaAutomatic,hu2013ppsgen,shibata2005automatic,inproceedings}. 
These studies mainly rely on a general assumption that a slide region is a form of summarization of the associated paper. Therefore, a wide range of summarization methods have been proposed to improve the effectiveness of the generation of slides.
As the next step for this research line, automatic design assistance for presentation slides can help the users have more effective presentations.

In a different context, some studies provide presenters with guidelines or alternatives to the traditional designs to communicate the presentation's content more effectively \cite{alley2004rethinking,Jennings:09,alley2006design,atkinson2005beyond,doumont2005slides}. 
These studies are in support of creating slides that feature sentence headlines and visual evidence to reinforce ideas and increase the audience’s retention of the information during presentation.

Emphasis Selection for written text in visual media was firstly proposed by \cite{shirani-etal-2019-learning}. An end-to-end sequence tagging architecture was proposed, which utilizes label distribution learning (LDL) to handle the subjectivity of the task and predict emphasis scores on short written texts. 
In a more recent study, 31 teams participated in SemEval’s 2020, \cite{shirani-etal-2020-semeval} and proposed different novel approaches to model emphasis effectively. 
For training and evaluation purposes, an emphasis selection dataset with social media short texts from Adobe Spark and publicly available quotes was introduced.
Top-performing teams, ERNIE \cite{huang2020ernie}, Hitachi \cite{morio2020hitachi}, and IITK \cite{singhal2020iitk} were able to achieve the first, second, and third places respectively by utilizing rich contextualized pre-trained language models such as 
ERNIE 2.0 \cite{sun2020ernie}, XLMRoBERTa \cite{conneau2019unsupervised}, XLNet \cite{NEURIPS2019_dc6a7e65}, and T5 \cite{raffel2019exploring}. 
In this study, we focus on a new domain, presentation slide, which emphasizes the importance of utilizing visual tools to convey a more effective presentation. 
Our work is quite different from prior works focusing on social media data as emphasis in presentation slides has different usage purposes in areas such as e-learning and marketing. Therefore, identifying emphasis in presentations brings unique challenges due to the differences in topic, length, and slide structure.

\section{Task Definition}
\label{sec:task}
Given a sequence of tokens in a slide page, $C=\{x_1,...,x_n\}$, the task is to compute a real value $y_i\in[0,1]$ for each $x_i$ in $C$, indicating the degree to which the token needs to be emphasized. 

\section{Data Collection}
\label{sec:datacolect}
The Presentation Slides Emphasis Dataset (PSED)\footnote{The dataset along with the annotations can be
found online: \url{https://github.com/RiTUAL-UH/Predicting-Emphasis-in-Presentation-Slides-Shared-Task}} is a collection of presentation slides, covering a wide range of topics from technical slides to non-technical ones such as children's material. 
Each instance of PSED represents one slide page along with eight annotations. 
To cover a wide range of topics and areas, we collected data from different sources such as websites with .ORG and .GOV domains and slides available on ACL anthology\footnote{\url{https://www.aclweb.org/anthology/}}. 
Since the slides are in different forms and shapes, we needed to apply some pre-processing steps to make sure the slide pages include clean pieces of text. Therefore, we removed slides that only contain equations, mathematical formulas, tables, or figures. 
To extract text and transcribe slide pages to written texts, we employed a python library called Pdfminer\footnote{\url{https://github.com/pdfminer/pdfminer.six}}.
We followed some quality control steps to ensure the text and the slide match together.

\subsection{Annotation Process}
In an MTurk experiment, we asked nine annotators to label each sample text by selecting word(s) on a slide page. 
More specifically, we showed the image of the slides as well as the corresponding raw texts. Workers were asked to select words that need to be emphasized as if they prepare the slides for their own presentation. 
To monitor the labeling process's quality, we included carefully-designed quality questions in 10 percent of the hits to make sure the annotators read the slides.

We observed a low Fleiss' Kappa score \cite{shrout1979intraclass} of 0.1414 on the dataset.
With a closer examination, we noticed some technical and domain-specific slides exist in the dataset that are not entirely understandable for the general audience. Therefore, we removed slides with Fleiss' Kappa score lower than -0.05. As a result, the overall Fleiss' Kappa score increased to 0.1797. We also noticed that in many cases, there is at least one annotator with a very different sets of selection. So to improve the agreement in the dataset and help to have a better training, for each slide, we identified and removed the annotator with the lowest agreement to the rest of the annotators. 
So the final dataset contains annotation from eight annotators and a Fleiss' Kappa score of 0.2092. Such a Kappa score indicates the existence of multiple points of view about emphasis in the dataset. Table \ref{tab:annotation-scheme} shows an example of a bullet point annotated with the BIO annotations. As it is shown, there are more agreements in selecting words such as ``risk" and ``management" compared to the rest. 

\begin{table*}[h]
\centering
\caption{An example of the collected data along with its eight annotations and emphasis probabilities. In the table, ``B"'s indicate the beginning of the emphasis and ``I"s indicate the inside and ``O"s indicate the non-emphasis words. ``Freq." column represents the frequency of ``B"s, ``I"s and ``O"s. The last column, ``Emphasis Probs.", shows the emphasis probability (``B+I") over eight annotations.}
\label{tab:annotation-scheme}
\begin{tabular}{c|cccccccc|c|c}
\toprule
Words       & A1 & A2 & A3 & A4 & A5 & A6 & A7 & A8 & Freq. [B,I,O]      & Emphasis Probs. [B+I] \\ \midrule\midrule
•           & O  & O  & O  & O  & O  & O  & O  & O  & {[}0,0,8{]} & 0.0  \\
Demonstrate & O  & B  & O  & O  & O  & O  & O  & O  & {[}1,0,7{]} & 0.125 \\
how         & O  & O  & O  & O  & O  & O  & O  & O  & {[}0,0,8{]} & 0.0  \\
operational & O  & O  & O  & O  & O  & O  & B  & O  & {[}1,0,7{]} & 0.125 \\
agencies    & O  & O  & O  & B  & O  & O  & O  & O  & {[}1,0,7{]} & 0.125 \\
are         & O  & O  & O  & O  & O  & O  & O  & O  & {[}0,0,8{]} & 0.0  \\
using       & O  & O  & O  & O  & O  & O  & O  & O  & {[}0,0,8{]} & 0.0  \\
NASA        & O  & B  & O  & O  & B  & O  & O  & O  & {[}2,0,6{]} & 0.25  \\
data        & O  & O  & O  & O  & I  & O  & O  & O  & {[}0,1,7{]} & 0.125 \\
for         & O  & O  & O  & O  & O  & O  & O  & O  & {[}0,0,8{]} & 0.0  \\
risk        & O  & O  & B  & O  & O  & O  & B  & B  & {[}3,0,5{]} & 0.375 \\
management  & B  & O  & I  & B  & O  & B  & I  & I  & {[}3,3,2{]} & 0.75 \\
\bottomrule
\end{tabular}
\end{table*}

\section{Data Analysis}
\label{sec:dataanalysis}
Table \ref{tab:stats2} provides details about the length of instances in PSED datasets. The table describes the minimum, mean, and maximum number of words in slides for each split.
The dataset contains 1,776 high-quality slides, randomly split up between (70\%) training, (10\%) development, and (20\%) test sets for further analysis. More information on the number of slides, sentences, and words in PSED dataset are provided in Table \ref{tab:stat1}.
\begin{table}[h]
\centering
 \caption{Statistics on the lengths of the samples}
 \label{tab:stats2}
 \begin{tabular}{c|c c c}
 \toprule
 Section & Min & Mean & Max\\
 \midrule\midrule
 Train  & 13 & 78 & 180 \\
 Dev  & 15 & 71 & 164 \\
 Test  & 17 & 79 & 181 \\
 \bottomrule
 \end{tabular}
\end{table}

\begin{table}[h!]
\centering
\caption{Dataset Statistics}
\begin{tabular}{c| c c c} 
 \toprule
 Section & \#Slides & \#Sentences & \#Words \\ [0.5ex] 
 \midrule\midrule
 Train & 1241 & 9645 & 96934 \\ 
 Dev & 180 & 1251 & 12822 \\
 Test & 355 & 2754 & 28108 \\
 Total & 1776 & 13650 & 137864 \\
\bottomrule
\end{tabular}
\label{tab:stat1}
\end{table}

\begin{figure}[h!]
  \centering
  \includegraphics[width=0.45\textwidth]{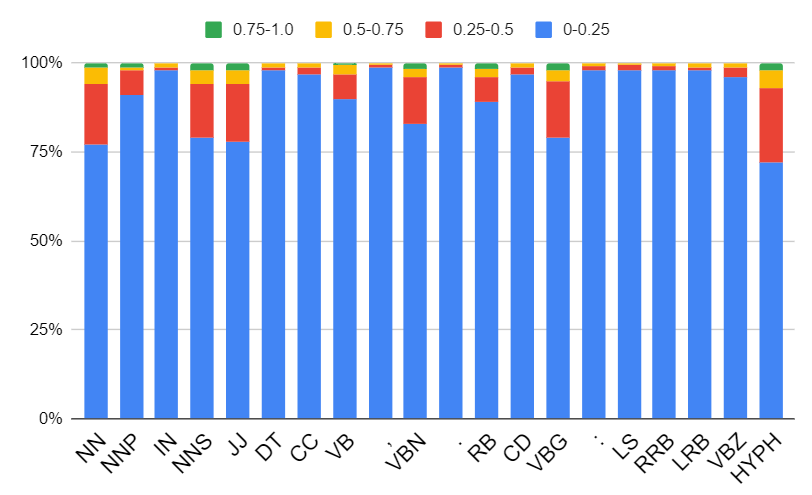}
  \caption{Frequencies of the top 20 POS tags in 0-0.25, 0.25-0.5, 0.5-0.75, 0.75-1.00 intervals of emphasis probabilities. The vertical values correspond to the percentage of tag counts over the total number of words in the training and development sets.}
  \label{fig:pos}
\end{figure}

\begin{figure}[ht]
\centering
\includegraphics[width=1\linewidth, right]{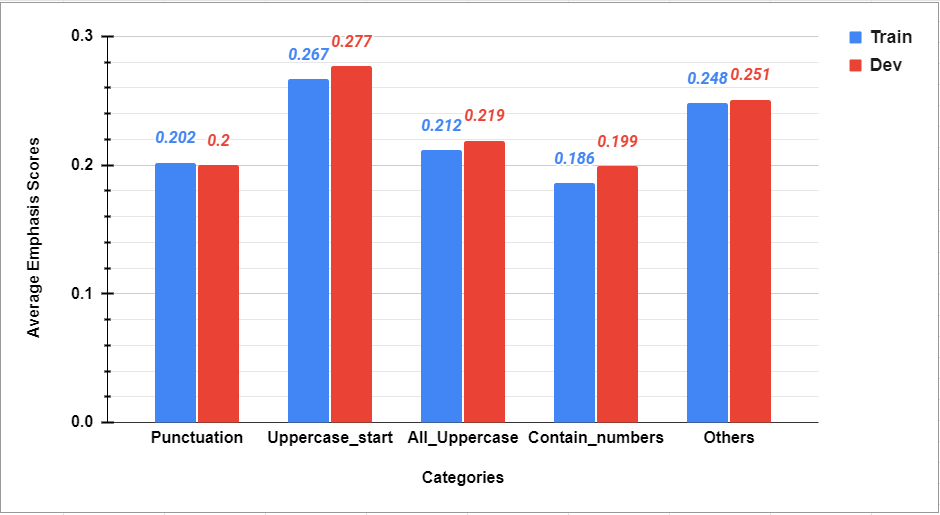}
\caption{The figure shows average emphasis scores on the training and development sets for four different lexical features. }
\label{fig:averave_emphasis_scores}
\end{figure}

Many systems reported performance improvement through employing Part-of-Speech-Tags (POS) as features to their models. In this section, we choose the top 20 POS tags, which frequently occur in the training and development sets, to analyze the feature's effectiveness. We used spaCy library \footnote{\url{https://spacy.io/usage/linguistic-features}} as a Part-of-speech tagger
to obtain POS tags for all tokens. To closely examine how the emphasis probabilities are distributed, we divided them into four intervals (0-0.25, 0.25-0.50, 0.50-0.75, and 0.75-1.00).
Figure \ref{fig:pos} shows the occurrence of the top 20 POS tags in four emphasis probability intervals for all token labels in our training and development sets. POS tags such as  ``IN", ``,", ``.", and ``:" are more favored to have low emphasis probabilities (0-0.25). Interestingly, some POS tags like ``DT", ``CD," and ``VBZ" have zero words in the highest emphasis probability interval (0.75-1.0). Overall, most tokens fall into the lowest emphasis probability, and the difference lies in (0.25-0.5) interval, where POS tags like ``NN", ``NNS," and ``VBG" mostly fall into.
Similar to POS tags, other hand-crafted features such as punctuations or uppercased tokens helped improve the results of some models. This motivates us to examine the degree of emphasis probability for different lexical features. 
Figure \ref{fig:averave_emphasis_scores} shows the average emphasis scores for each category in the training and development sets. 
Comparing all lexical features, ``UpperCase\_start" has the highest average emphasis score, and ``Contain\_numbers" and ``Punctuations" are more favored to have the lowest average scores. This indicates some general trends for emphasized words in the slides with respect to the categories of words.


\section{CAD21 Shared Task}
To better examine the challenges of the dataset and benchmark the task, we organized a shared task and invited the community to participate in modeling emphasis in a new challenging domain. 
There is a total of four teams participating in CAD21 shared task during the evaluation phase. We observed many novel and interesting set of solutions for this particular task, from non-transformer-based models such as BiLSTM-ELMo to more advanced pre-trained models such as XLNet, RoBERTa, ERNIE 2.0 and SciBERT. 
The most popular approach was ensemble Transformer-based models. Many hand-crafted features such as Part-of-speech (POS) tags, keywords, and lexical features (such as words with capital letters and punctuation) were explored to improve the models' performance.

\begin{table*}[h]
\centering
\caption{List of teams that participated in CAD21 Shared Task with their ranks and score}
\label{tab:leaderboard}
\begin{tabular}{c|cccc}
\toprule
Teams      & \textit{RANK}  & Score 1   & Score 5   & Score 10  \\ \midrule\midrule
UBRI-604   & \textbf{0.525} & \textbf{0.335} (1) & \textbf{0.686} (1) & 0.554 (2) \\
DeepBlueAI & 0.519 & 0.330 (2) & 0.667 (3) & \textbf{0.559} (1) \\
Cisco      & 0.518 & 0.330 (2) & 0.675 (2) & 0.551 (3) \\
Baseline   & 0.475 & 0.301 (3) & 0.634 (5) & 0.489 (5) \\
Zouwuhe   & 0.474 & 0.285 (4) & 0.638 (4) & 0.500 (4) \\
\bottomrule
\end{tabular}
\end{table*}
\begin{table*}[h!]
\centering
\caption{Length vs. Performance on the test set }
\label{tab:length_vs_performace}
\begin{tabular}{@{}l|llll@{}}
\toprule 
Length/Performance & \textit{RANK} & Score 1 & Score 5 & Score 10 \\ \midrule\midrule
Short (\textless 60 tokens, 112 slides) & \textbf{0.601} & \textbf{0.42} (1) & \textbf{0.634} (1) & \textbf{0.75} (1) \\
Medium (60 - 90 tokens, 126 slides) & 0.55 & 0.349 & 0.589 & 0.713 \\
Long (\textgreater 90 tokens, 116 slides) & 0.485 & 0.293 & 0.526 & 0.635 \\ \bottomrule
\end{tabular}
\end{table*}

\subsection{Evaluation Metric}
\label{sec:metric}
We followed a similar evaluation method used in \cite{shirani-etal-2020-semeval}. We compute $\mbox{Match}_m$ metric for 1, 5, and 10 words with top probabilities on the test set. This metric is specifically designed to meet the subjectivity of the task. 

\paragraph{$\mbox{Match}_m$} For each slide page $x$ in the test set $D_{test}$, we select a set $S^{(x)}_m$ of $m \in \{1,5,10\}$ words with the top $m$ probabilities according to the ground truth. Similarly, we select a prediction set $\hat{S}^{(x)}_m$ for each $m \in \{1,5,10\}$, based on the prediction probabilities.
The $\mbox{Match}_m$ is defined as follows:
\[\mbox{Match}_m := \frac{\sum_{x \in D_{test}} |S^{(x)}_m \cap \hat{S}^{(x)}_m|/m}{|D_{test}|}\]
In order to better compare the results, we compute the average value of $\mbox{Match}_m$ for all $m \in \{1,5,10\}$ and we call this averaged value (\textit{RANK}). 
We treat words in the ground truth with the same probability equally, so if the model predicts either of the tokens, we consider it as a correct answer. 

\subsection{Baseline Model}
\label{sec:baseline}
In this section, we discuss the baseline model used for this task. To demonstrate the challenges of this task compared to \cite{shirani-etal-2020-semeval}, we decided to use the same baseline model (DL-BiLSTM-ELMo), which is introduced in \cite{shirani-etal-2019-learning}. 
With a sequence-labeling architecture, this model utilizes ELMo contextualized embeddings \cite{peters2018deep} as well as two BiLSTM layers to label emphasis. Moreover, the Kullback-Leibler Divergence (KL-DIV) ~\cite{kullback1951information} is used as the loss function during the training phase.

\subsection{Systems and Results}
\label{sec:results}
Four teams participated in CAD21 shared task during the evaluation phase. The results of the four scores, as well as the \textit{RANK} score, are shown in Table \ref{tab:leaderboard}. \emph{UBRI-604} \cite{hu2021CAD21} and \emph{Cisco} \cite{ghosh2021CAD21} submitted system description papers including their data analysis, detailed experiments and reported results. In total, three teams performed higher than the baseline, and one team performed lower. 

The top-performing team, \emph{UBRI-604} \cite{hu2021CAD21}, ranked in the first place with \textit{RANK} score of (0.525) on the leader board 
in the evaluation phase. \emph{UBRI-604} achieved the highest score on both Score 1 and Score 5. 
\emph{DeepBlueAI} team stood in second place (0.519), with a 0.006 \textit{RANK} score lower than the first team. 
\emph{DeepBlueAI} outperformed all teams in Score 10 by achieving 0.559.
Finally, \emph{Cisco} \cite{ghosh2021CAD21}, with 0.001 scores lower than the second team ranked third. 
In the next section, we describe and compare the approaches the three teams used to model emphasis.

\subsection{Top Performing Systems and Novel Architectures}
\label{sec:architectures}
\emph{UBRI-604} \cite{hu2021CAD21} proposed an end-to-end Transformer-based 
approach. 
Different rich Transformer-based pre-trained language models were explored during the experiment, such as 
ALBERT \cite{lample2019cross}, GPT-2 \cite{radford2019rewon}, ROBERTA \cite{liu2019roberta}, ERNIE 2.0 \cite{sun2020ernie}, XLNET\cite{NEURIPS2019_dc6a7e65}, XLM-ROBERTA and
BERT \cite{devlin2019bert}. Comparing the results of all seven models, XLM-ROBERTA-LARGE performed the best. 
Besides pre-trained language models, \emph{UBRI-604} explored hand-crafted features. Their model leveraged the lexical features (such as words with capital letters and punctuations) for further improvement.

The second team, \emph{DeepBlueAI}, introduced an ensemble Transformer-based model with two fully-connected layers combined with POS tags embedding and hand-crafted features. 
The ensemble model takes advantage of BERT, SciBERT \cite{beltagy2019scibert} and ERNIE 2.0 pre-trained language models 
by taking the average of the scores predicted by these models.

\emph{Cisco} \cite{ghosh2021CAD21} explored two approaches based on BiLSTM+ELMo \cite{shirani-etal-2019-learning} architecture and Transformer-based pre-trained models with the base model of RoBERTa and XLNET. 
They enriched the ELMo contextual embedding in BiLSTM+ELMo model by incorporating a character-level BiLSTM Network. Interestingly, the reported results show an increase of 0.026 when POS tags and keyphrases are added to the model. This shows the effectiveness of these two features for this task.
\emph{Cisco's} best score on the evaluation phase was obtained using an ensemble of XLNet and RoBERTa that makes them the third in the leaderboard. They boosted the model further in the Post Evaluation phase by ensembling XLNet and BiLSTM-ELMo models and incorporating hand-crafted features like POS and Keyphrase. 
In an interesting analysis, they showed the model's performance deteriorates with the slides' increasing length. 


\section{Discussion}
\label{sec:Discussion}

PSED dataset contains slides with different lengths. To better examine how the length of slides can affect the prediction, we perform an error analysis to examine this relationship. 
We divided the test set into three sections based on the instances length, namely \textless 60, 60-90, and \textgreater 90 tokens. Then we compute the average $\mbox{Match}_m$ scores over all submissions for every example in each section.
As shown in Table \ref{tab:length_vs_performace}, short slides always achieve better scores compared to the longer ones (medium and long slides) across all score 1, 5, and 10.

Many slides in PSED dataset contain scientific words. Besides using pre-trained models, trained on a general domain, some teams decided to handle scientific words differently. For example, \emph{DeepBlueAI} explored using the SciBERT \cite{beltagy2019scibert} model, which is pre-trained on scientific articles. On the other hand, \emph{Cisco} explored training a scientific keywords predictor and use the output as a feature to the model. 
Extending the proposed approaches to more efficiently address the diverse vocabulary of the dataset is an important future direction.

\section{Conclusion}
\label{sec:Conclusion}
We have presented a new dataset for emphasis selection on presentation slides. The dataset poses new challenges for modeling emphasis. As part of this study, we set out a shared task and invited researchers to model emphasis for the first time on the domain of presentation slides. 
We provided different analysis on the dataset and summarized the insights gained from the shared task. A future extension could explore more robust techniques to address the challenges in PSED dataset with a large number of slides diverse in topic, structure and length.





 
\bibliographystyle{aaai}
\bibliography{aaai.bib}
 
\end{document}